\def\year{2022}\relax
\documentclass[letterpaper]{article} 
\usepackage{aaai22}  
\usepackage{times}  
\usepackage{helvet}  
\usepackage{courier}  
\usepackage[hyphens]{url}  
\usepackage{graphicx} 
\usepackage{natbib}  
\usepackage{caption} 
\DeclareCaptionStyle{ruled}{labelfont=normalfont,labelsep=colon,strut=off} 
\usepackage{algorithm}
\usepackage{algorithmic}
\usepackage{amsmath}
\usepackage{xcolor}

\usepackage{newfloat}
\usepackage{listings}
\floatstyle{ruled}
\newfloat{listing}{tb}{lst}{}
\floatname{listing}{Listing}
\title{2nd Place Solution for ICCV2021 VIPriors Image Classification Challenge: \\
An Attract-and-Repulse Learning Approach}
\author{
    Yilu Guo\textsuperscript{\rm 1}\equalcontrib,
    Shicai Yang\textsuperscript{\rm 1,2}\equalcontrib,
    Weijie Chen\textsuperscript{\rm 1,2}, \\
    Liang Ma\textsuperscript{\rm 1}, 
    Di Xie\textsuperscript{\rm 1},
    Shiliang Pu\textsuperscript{\rm 1}\thanks{Corresponding author.}
}

\affiliations{
    \textsuperscript{\rm 1}Hikvision Research Institute, Hangzhou, China\\
    \textsuperscript{\rm 2}Zhejiang University, Hangzhou, China\\
    \{guoyilu5, yangshicai, chenweijie5, maliang6, xiedi, pushiliang.hri\}@hikvision.com
}

\usepackage{bibentry}

\begin{document}

\maketitle

\begin{abstract}
Convolutional neural networks (CNNs) have achieved significant success in image classification by utilizing large-scale datasets. However, it is still of great challenge to learn from scratch on small-scale datasets efficiently and effectively. With limited training datasets, the concepts of categories will be ambiguous since the over-parameterized CNNs tend to simply memorize the dataset, leading to poor generalization capacity. Therefore, it is crucial to study how to learn more discriminative representations while avoiding over-fitting. Since the concepts of categories tend to be ambiguous, it is important to catch more individual-wise information. Thus, we propose a new framework, termed Attract-and-Repulse, which consists of Contrastive Regularization (CR) to enrich the feature representations, Symmetric Cross Entropy (SCE) to balance the fitting for different classes and Mean Teacher to calibrate label information. Specifically, SCE and CR learn discriminative representations while alleviating over-fitting by the adaptive trade-off between the information of classes (attract) and instances (repulse). After that, Mean Teacher is used to further improve the performance via calibrating more accurate soft pseudo labels. Sufficient experiments validate the effectiveness of the Attract-and-Repulse framework. Together with other strategies, such as aggressive data augmentation, TenCrop inference, and models ensembling, we achieve the second place in ICCV 2021 VIPriors Image Classification Challenge.
\end{abstract}

\section{Introduction}

Convolutional neural networks (CNNs) have achieved tremendous success in image classification. However, it deeply depends on large-scale datasets, such as ImageNet~\cite{deng2009imagenet} and OpenImage~\cite{kuznetsova2018the}. Generally, CNNs learn to generalize well with massive data. When trained on a small-scale dataset, they are required to be pre-trained on a large-scale dataset in a supervised or unsupervised manner. Herein, we can't help to ask, can we achieve comparable results on a small dataset by learning from scratch without any pre-training? This is an interesting and significant topic proposed in the VIPriors Image Classification Challenge~\cite{bruintjes2021vipriors}.

CNNs are data-hungry and they usually summarize different feature patterns by learning from large-scale dataset. While when the training data is limited, especially when the image amount of each category is quite small, the concept of category tends to be ambiguous. It is hard for CNNs to generalize the concept of category by merely exploiting a small amount of samples, but CNNs will still memorize the data resulting in over-fitting. Hence, it is a challenging problem to extract discriminative representations by learning from scratch on a very small-scale dataset. In addition, it is crucial to alleviate over-fitting since the models with a large network capacity are prone to memorize the dataset, leading to a poor generalization ability. 

Recently, self-supervised learning has shown a great potential of learning discriminative representation from the training data without external label information. And the representations from self-supervised learning exhibit better generalizability than the counterparts from supervised learning~\cite{tendle2021a,sariyildiz2021concept}. In particular, the contrastive learning methods~\cite{chen2020a,he2020momentum} have demonstrated advantages over other self-supervised learning methods in learning better transferable representations for downstream tasks, like object detection and semantic segmentation. Contrastive learning is a remarkable self-supervised learning framework that learns invariant features by contrasting positive samples against negative ones without annotations. The representations learned by contrastive learning are individually discriminative and unbiased to the image labels, which can effectively prevent the model from over-fitting the classification patterns of any object category.

\begin{figure*}[h]
\centering
\includegraphics[width=6.5in]{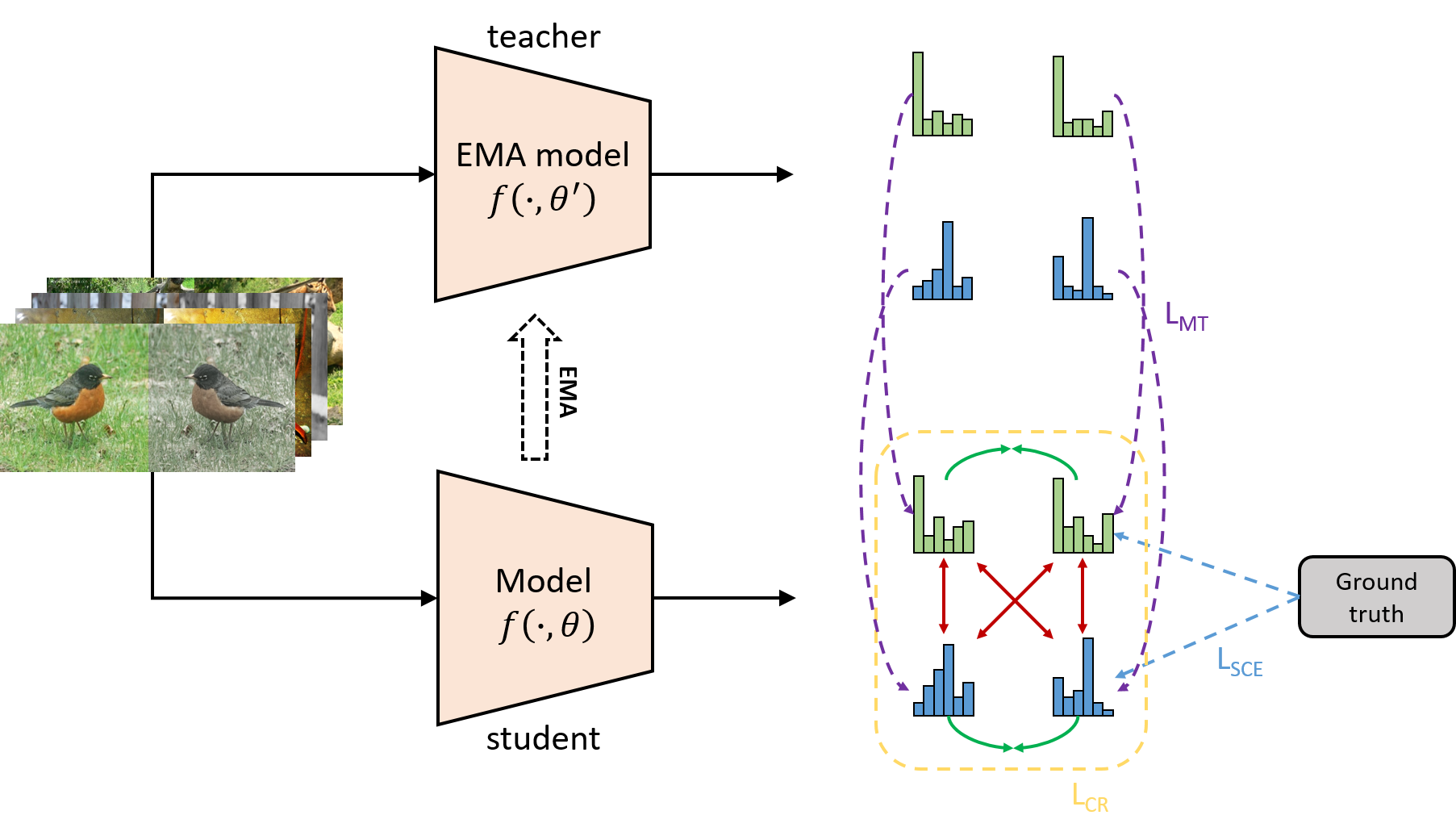}
\vskip -0.1in
\caption{The pipeline of Attract-and-Repulse. The two views of images by different augmentation are packed and fed to both teacher and student models. For simplicity, $\delta$ is set 0 here. The views of the same images are regarded as positive pairs (\textbf{\textcolor{green}{green}} lines) while the negative ones (\textbf{\textcolor{red}{red}} lines) are formed by different images within a mini-batch in the Contrastive Regularization (CR). The \textbf{\textcolor{yellow}{yellow}} dotted box indicates the CR loss, the colored distributions indicate the outputs of models, and the distributions for the same image are denoted by the same color. The Symmetric Cross Entropy (SCE) can balance the fitting of different classes. After learning by CR and SCE, the exponential moving average (EMA) model (Mean Teacher) can produce a more accurate soft pseudo label for each augmented images to optimize the student model.}
\label{fig:pipeline}
\end{figure*}

In this paper, we aim to explore the capability of contrastive learning under the data-deficient setting. We introduce contrastive learning into the class-probability space predicted by the model to strengthen the feature representations by individual discrimination and avoid over-fitting the ambiguous categories, which is termed as Contrastive Regularization (CR). And when the training data is deficient, the learning of different classes is prone to be unbalanced, we further use Symmetric Cross Entropy (SCE)~\cite{wang2019symmetric} to balance the fitting of different classes. The SCE makes different classes to attract instances more evenly while the CR lets instances repulse with each other. The model can learn discriminative representations while alleviating over-fitting by the adaptive trade-off between attraction and repulsion. After attracting and repulsing, instances can reach a suitable position among different classes, and Mean Teacher~\cite{tarvainen2017mean} is used to improve the performance with the more accurate soft pseudo label. In conclusion, the Contrastive Regularization, Symmetric Cross Entropy, and Mean Teacher compose our proposed Attract-and-Repulse framework. Together with other strategies, such as aggressive data augmentation, TenCrop inference, and models ensembling, the proposed Attract-and-Repulse framework achieves a very competitive performance in the VIPriors Image Classification Challenge~\footnote{We win the second place in ICCV 2021 VIPriors Image Classification Challenge. See Section 2.1.2 in the tech report \cite{DBLP:journals/corr/abs-2201-08625}}. It is worth emphasizing that we achieve 74.49\% top-1 accuracy on ImageNet~\cite{krizhevsky2017imagenet} by using only about 8\% training data, which surpasses the previous state-of-the-art approaches by a large margin.

To summarize, our main contributions are highlighted as:
\begin{itemize}
\item We propose Contrastive Regularization to strengthen the feature representations by individual discrimination and avoid over-fitting the ambiguous categories representations in the data-deficient setting.
\item We propose Attract-and-Repulse, a novel framework to deal with data-deficient image classification, which learns discriminative representations while alleviating over-fitting by the adaptive trade-off between the information of classes (attract) and instances (repulse).
\item Together with augmentation strategies and other strategies, Attract-and-Repulse achieves competitive performance in the VIPriors Image Classification Challenge.
\end{itemize}

\section{Related Works}
\subsection{Contrastive Learning}
Recently, contrastive learning approaches~\cite{chen2020a,he2020momentum,grill2020bootstrap} emerge as the mainstream paradigm in self-supervised learning which learns invariant features by contrasting positive samples against negative ones. A positive pair is usually formed with different augmented views of the same image, while negative ones are formed with different images. Particularly, SimCLR ~\cite{chen2020a} obtains positive and negative pairs within a mini-batch of training data and uses InfoNCE ~\cite{oord2018representation} loss to learn the feature representations. It requires a large batch size to effectively balance the positive and negative ones. MoCo~\cite{he2020momentum} builds a large and consistent feature queue to store negative samples using a slowly progressing momentum network which greatly reduces high memory cost. BYOL ~\cite{grill2020bootstrap} and SimSiam ~\cite{chen2021exploring} challenge the indispensability of negative examples and achieve impressive performance by only using positive ones. Before BYOL and SiaSiam, UIC~\cite{Unsupervised2020} can be viewed as an earlier one which uses positive samples for contrastive learning without negative samples. SwAV ~\cite{caron2020unsupervised} obtains a better performance by enforcing consistent cluster assignment prediction between multiple views of the same image. Supervised Contrastive Learning (SCL)~\cite{khosla2020supervised} adapts contrastive learning to the fully supervised setting to learn more informative representations by effectively leveraging label information.
Furthermore, contrastive learning has promoted the performance of various tasks, including semi-supervised learning~\cite{chen2020big,li2020comatch}, learning with noisy label~\cite{zheltonozhskii2021contrast,Friends} and so on.

\subsection{Data Augmentation}
Data augmentation is an effective way to improve CNNs’ generalization performance especially in the case of insufficient data. Mixup~\cite{zhang2017mixup} trains a model on elementwise convex combinations of pairs of examples and their labels together. Cutout~\cite{devries2017improved} and random erasing~\cite{zhong2020random} randomly erase rectangle regions on input images during training. Rather than occluding a portion of an image, CutMix~\cite{yun2019cutmix} replaces a patch of an image with a patch of a different image where the training labels are also mixed proportionally to the area of patches. Recently, with the emergence of AutoML, network learning strategies also can be searched from data. AutoAugmentation~\cite{cubuk2019autoaugment} originally uses reinforcement learning to choose a sequence of operations as well as their probability of application and magnitude. Since AutoAugmentation needs a huge space for searching, RandAugmentation~\cite{cubuk2020randaugment} proposes a simplified search space that has less computational expense. TA$^3$~\cite{Target2022} utilizes RandAugmentation to enhance unsupervised domain adaptive object detection.

\section{Method}
Attract-and-Repulse is a new framework for data deficient learning, which consists of Contrastive Regularization, Symmetric Cross Entropy, and Mean Teacher. The entire pipeline is shown in Figure~\ref{fig:pipeline}.

\subsection{Preliminaries}
Given a $C$-class dataset $D=\{(x_i,y_i)\}_{i=1}^n$, with $x\in\mathcal{X}\subset {R}^{d}$ denoting a sample in the $d$-dimensional input space and $y\in\mathcal{Y}=\{1,\cdots,C\}$ its associated label. In the data-deficient setting, the number of samples per class is quite small. 
For each sample $x$, a classifier $f(x)$ computes its probability of each label $k\in\{1,\cdots,C\}:p(c|x)=\frac{e^{z_c}}{\sum_{j=1}^C e^{z_j}}$, where $z_j$ are the logits. We denote the ground-truth distribution over labels for sample $x$ by $q(c|x)$, and $\sum_{c=1}^k q(c|x)=1$. Consider the case of a single ground-truth label $y$, then $q(y|x)=1$ and $q(c|x)=0$ for all $c\neq y$. The cross entropy loss for sample $x$ is (denote cross entropy as $H(q,p)$):
$$L_{ce}=H(q,p)=-\sum_{c=1}^K q(c|x)\log{p(c|x)}$$

\subsection{Contrastive Regularization}

\begin{figure}
  \centering
  \includegraphics[scale=0.25]{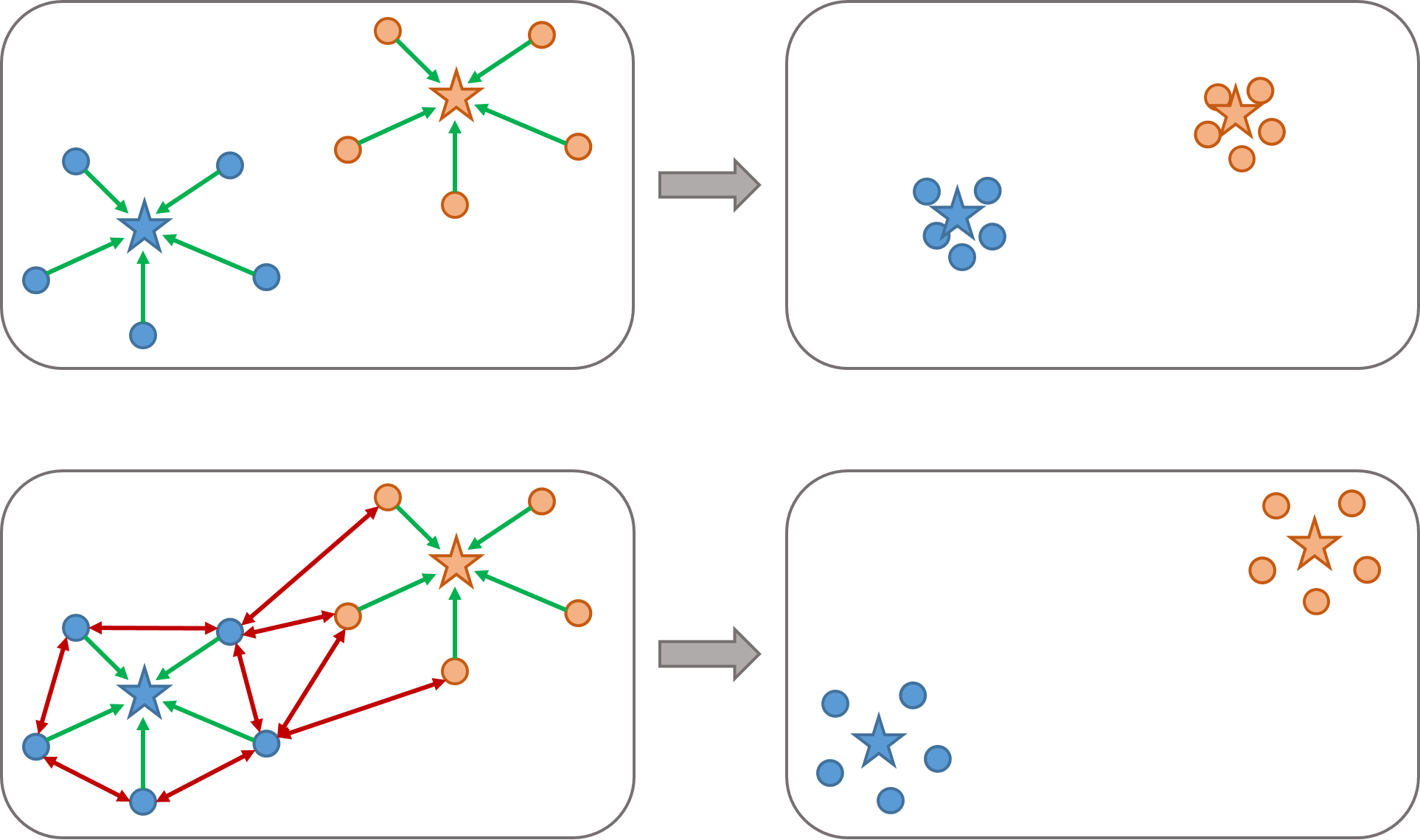}
  \caption{Diagram for the learning process using Cross Entropy (top) and Cross Entropy with Contrastive Regularization (bottom). The stars and the circles indicate the class centers and the instances respectively while the same class are denoted by the same color. The \textbf{\textcolor{green}{green}} lines and the \textbf{\textcolor{red}{red}} lines indicate the attraction and the repulsion, respectively. For simplicity, we don't draw all red lines.}
  \label{fig:ce_cr}
\end{figure}

When the data is deficient, the concepts of some categories will be more ambiguous. While the cross entropy loss attracts samples close to their own class centers, as shown on the top of Figure~\ref{fig:ce_cr}, the model will be easy to over-fit the ambiguous classes and resulting in weak generalization ability. Inspired by the appearance that the differences among the samples in an ambiguous class are larger than that in explicit one, we adapt contrastive learning to the model learning, which is called Contrastive Regularization. Contrastive Regularization makes the samples repulse with each other in the same classes and forms confrontation with the attraction of class centers, as shown on the bottom of Figure~\ref{fig:ce_cr}. The trade-off between attraction and repulsion prevents the model from over-fitting the ambiguous classes. What's more, the repulsion among instances from the same class can make the instances disperse more uniformly around the class center, while the repulsion among instances from different classes can increase the distance between classes. And Contrastive Regularization also introduces more individual information into model learning, while catching more information is important in the data-deficient setting. 

Supervised Contrastive Learning (SCL)~\cite{khosla2020supervised} adapts contrastive learning to the fully supervised setting to learn more informative representations by effectively leveraging label information. SCL is an excellent representation learning method, but the model learned by SCL usually needs to finetune on downstream tasks. Contrastive Regularization integrate SCL with the supervised target task by applying SCL on the final prediction of the model.

Within a multiview batch, let $i\in I\equiv \{1...2N\}$ be the index of an arbitrary augmented smaple, the original loss of SCL is as follow:
$$L_{SCL}=\sum_{i\in I} {\frac{-1}{\vert{K(i)}\vert}} {\sum_{k\in K(i)} \log{\frac{\exp(z_i\cdot z_k/\tau)}{\sum_{a\in A(i)} \exp(z_i\cdot z_a/\tau)}}}$$
Here, $z_l$ is the normalized feature representation, $\tau$ is a temperature hyper-parameter, $A(i)\equiv I \setminus\{i\}$, $K(i)\equiv \{k\in A(i) : \tilde{\pmb{y}}_k = \tilde{\pmb{y}}_i\}$ is the set of indices of all positives in the multiviewed batch distinct from $i$, and $\vert{K(i)}\vert$ is its cardinality.

To make the confrontation between attraction and repulsion more direct, the proposed Contrastive Regularization changes the feature $z_l$ to the probability distribution $p_l$ of the model output:
$$L_{CR}=\sum_{i\in I} {\frac{-1}{\vert{K^{'}(i)}\vert}} {\sum_{k\in K^{'}(i)} \log{\frac{\exp(p_i\cdot p_k/\tau)}{\sum_{a\in A(i)} \exp(p_i\cdot p_a/\tau)}}}$$
Here, $K^{'}(i)\equiv \{k\in A(i) : dis(\tilde{\pmb{y}}_k, \tilde{\pmb{y}}_i)\leq\delta\}$. 
For alleviating the over-fitting problem, cutmix and mixup are usually used so that the hard-label are changed to soft-label and $K(i)$ is changed correspondingly. Kullback-Leibler divergence is usually used to measure the discrepancies between two distributions, while it is asymmetric. So we use Jensen-Shannon divergence, the symmetric variant of Kullback-Leibler divergence as the $dis$, and use $\delta$ to adjust the attention level to individual-wise or class-wise contrasting. The $\delta$ is smaller, the positive pairs need to have more similar soft-label distribution. Especially, when $\delta=0$ the CR degrades to vanilla InfoNCE that only regards the two views of one image by different augmentation as positive pairs. A small $\delta$ is used to strengthen the individual information to balance the fitting of the ambiguous classes.

\subsection{Symmetric Cross Entropy}

Symmetric Cross Entropy (SCE)~\cite{wang2019symmetric} is a simple yet effective loss for learning with noisy label. It aims to simultaneously address the hard class learning problem and the noisy label overfitting problem of Cross Entropy.

\begin{figure}[t]
\centering
\includegraphics[width=3.2in]{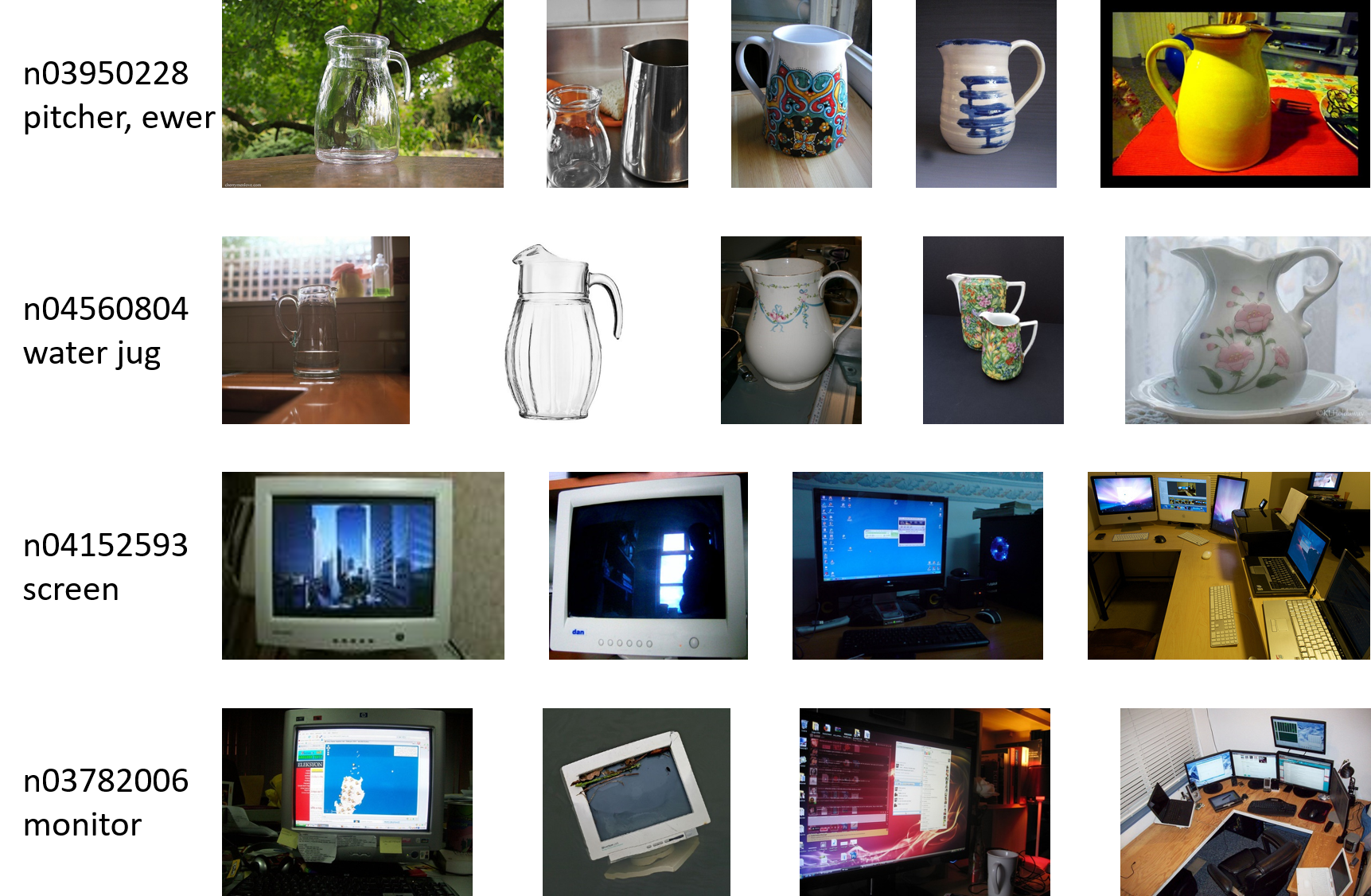}
\caption{Some examples for similar category concepts in Imagenet dataset. The corresponding labels are on the left side of the pictures. The images and labels in the top two rows are both very similar and as well as the bottom two.}
\label{fig:noisy}
\end{figure}

The label of ImageNet dataset is well-known to contain errors~\cite{northcutt2021confident,northcutt2021pervasive}. And there are many similar category concepts (see Figure~\ref{fig:noisy}) that will be more ambiguous when the image amount of each category is limited. So, there may be some ``noisy'' labels that may have a correct label but also can be deemed to another class. Thus, we employ Symmetric Cross Entropy~\cite{wang2019symmetric} to balance the fitting of different classes. The Symmetric Cross Entropy is easily constituted by standard cross entropy and reverse cross entropy.
$$L_{SCE}=L_{CE}+\alpha L_{RCE}=H(p,q)+\alpha H(q,p)$$

\begin{algorithm}[t]
\caption{Attract-and-Repulse learning framework}
\label{alg:algorithm}
\textbf{Input}: batch size $N$, total training step $T$, augmentation $\mathcal{A}$\\
\textbf{Parameter}: model weights $\theta$, ema model weights $\theta^{'}$ ($\theta_0^{'}=\theta_0$)
\begin{algorithmic}[0]
\WHILE{$t < T$}
\FOR{sampled minibatch $\{(x_i, y_i)\}_{i=1}^N$}
\FOR{all $i \in \{1,\cdots,N\}$}
\STATE draw two augmentation functions $a\sim \mathcal{A}$, $a^{'}\sim \mathcal{A}$
\STATE $v_i=a(x_i)$, $v_i^{'}=a^{'}(x_i)$ 
\ENDFOR
\STATE perform the same cutmix or mixup for 
$\{(v_i, y_i)\}_{i=1}^N$, $\{(v_i^{'}, y_i)\}_{i=1}^N$, respectively
\ENDFOR
\STATE $L=L_{CE}+\alpha L_{RCE}+\beta L_{CR}+\gamma L_{MT}$ \\
update $\theta$ by minimizing $L$
\STATE update $\theta^{'}$ by $\theta_t^{'}=\eta\theta_{t-1}^{'}+(1 - \eta)\theta_t$
\ENDWHILE
\STATE \textbf{return} $\theta_t$
\end{algorithmic}
\end{algorithm}
\subsection{Mean Teacher}
Mean Teacher~\cite{tarvainen2017mean} is proposed for semi-supervised learning. Here, we adapt it to provide stabilized and more accurate pseudo labels and stabilize model learning.

After learning by the CR and the SCE, the trade-off between individual-wise information and class-wise information makes the samples to be situated at a suitable position among different classes. So the model can produce more informative pseudo labels than the original labels. We maintain the exponential moving average (EMA) weights of the model and use the EMA weights as a teacher model to provide stabilized and accurate pseudo labels. Formally, we define $\theta_t^{'}$ at training step $t$ as the EMA of successive $\theta_t$ weights.
$$\theta_t^{'}=\eta\theta_{t-1}^{'}+(1 - \eta)\theta_t$$

And we define the consistency loss $L_{MT}$ as the expected Kullback-Leibler divergence between the prediction of the student model and the teacher model.
$$L_{MT} = {E}_x[KL(f(x,\theta)||f(x,\theta^{'}))]$$

Averaging model weights over training steps tends to produce a stabilized model and can provide more accurate soft pseudo labels. Thus, the model can avoid learning some inaccurate information via consistency loss and the learning process will be more robust.

\subsection{Attract-and-Repulse framework}
Finally, we reach a novel approach to deal with the data-deficient image classification, $i.e.$ Attract-and-Repulse framework. It optimizes the Symmetric Cross Entropy loss with the Contrastive Regularization to catch and balance the information from instances and classes. And adopting Mean Teacher to use the more accurate pseudo labels. Algorithm~\ref{alg:algorithm} summarizes the proposed framework and the overall loss function of Attract-and-Repulse can be formulated as follows:

\begin{equation}
    \begin{aligned}
    L &= L_{SCE}+\beta L_{CR}+\gamma L_{MT} \\
      &= L_{CE}+\alpha L_{RCE}+\beta L_{CR}+\gamma L_{MT} 
    \end{aligned}
    \label{E_L}
\end{equation}

\subsubsection{Auxiliary Classifier}

Supervisions to the intermediate output are usually used in deep learning to reduce the difficulty of optimizing the deep network~\cite{szegedy2016rethinking,zhao2017pyramid} or to enhance the information from different scales~\cite{xie2017holistically}. It is difficult to extract ample image information in the data-deficient setting, so we add an Auxiliary Classifier to the intermediate output of the model. And during inference, the prediction is computed by the weighted average of the intermediate output and the final output, which is termed as Auxiliary Fusion.

\section{Experiments}
\subsection{Dataset and Experimental Setting}
Visual Inductive Priors (VIPriors) Image Classification Challenge~\cite{bruintjes2021vipriors} proposes the topic that how to learn from scratch in a data-deficient setting. The objective of VIPriors Image Classification Challenge is to increase the Top-1 Accuracy on ImageNet dataset~\cite{deng2009imagenet} by only using a small subset of ImageNet dataset. The data is divided into three splits, including a training set, a validation set, and a testing set that is unavailable during the model optimization. The training and validation splits are two subsets of the original training split. The test set is taken from the original validation split directly. Each split includes 1,000 classes which are the same as the original ImageNet and 50 images per class, resulting in 50,000 images in total.

We train models on the subset of the ImageNet which was provided by the VIPrior Image Classification Challenge without any pre-trained models \footnote{https://github.com/VIPriors/vipriors-challenges-toolkit/tree/master/image-classification}. In the following section of ablation study, we mainly train models on the training split (about \textbf{4\%} training data of the original ImageNet) and verify on the validation split. And in the section of comparison with other state-of-the-arts, we combine the training and validation splits for training (about \textbf{8\%} training data of the original ImageNet) and randomly split a few samples for validation. 

\begin{figure}[t]
  \centering
  \includegraphics[scale=0.42]{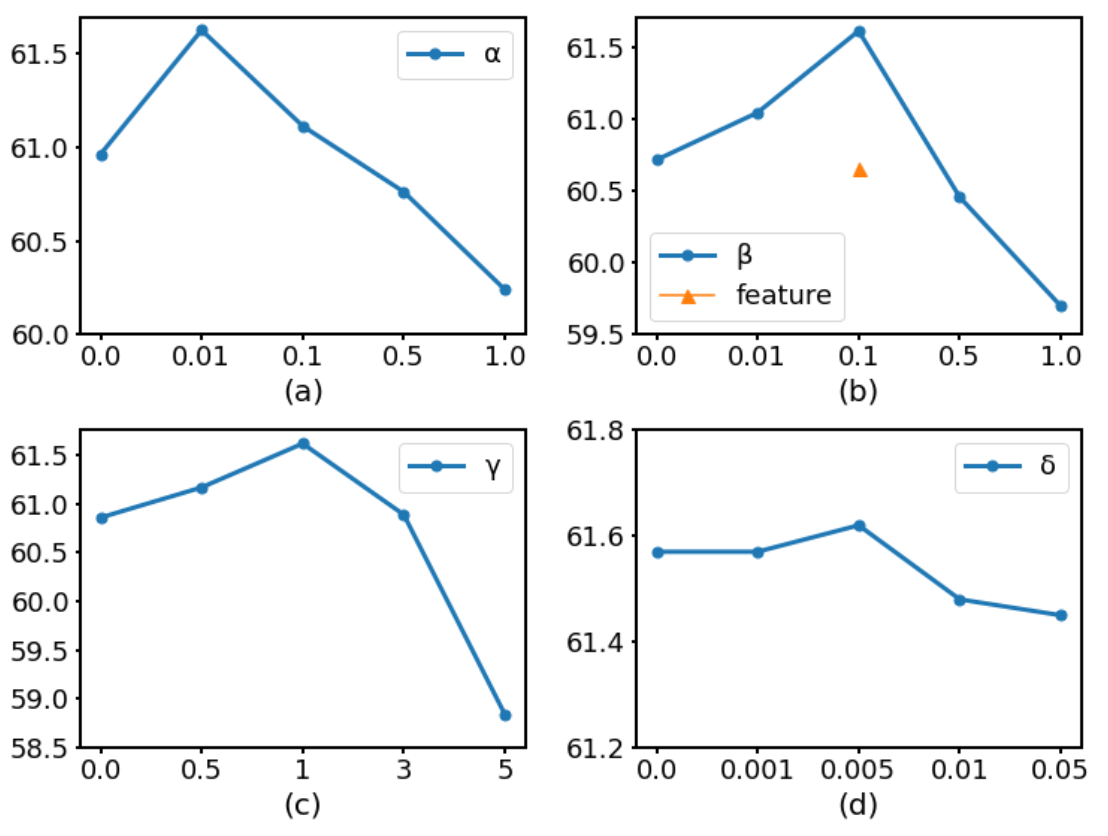}
  \caption{The effect of four hyper-parameters ($\alpha$, $\beta$, $\gamma$, and $\delta$). The y-axes are the top-1 accuracy (\%). The \textbf{\textcolor{orange}{orange}} triangle in the top right sub-figure denotes that we apply Contrastive Regularization in the feature space instead of the class-probability space. }
  \label{fig:ablation}
\end{figure}
\begin{table}[t]
\begin{center}
\begin{tabular}{ll}
\hline\noalign{\smallskip}
Model & top-1 Acc. (\%) \\
\noalign{\smallskip}
\hline
EfficientNet-b2 & 41.11 \\
EfficientNet-b4 & 49.40 \\
ResNet50 & 40.91 \\
ResNeXt101\_32x4d & 45.32 \\
Swin Transformer & 26.43 \\
\hline
\end{tabular}
\caption{Ablation study on model architecture. The transformer architecture ``Swin-Transformer'' is infeasible in the setting of data deficient without any pre-training.}
\vskip -0.1in
\label{table:model}
\end{center}
\end{table}

\subsection{Implementation Details}

We use the RMSprop~\cite{dauphin2015rmsprop} optimizer with alpha set to 0.9 and momentum set to 0.9. Models are trained with 8 GPUs and 64 samples per GPU. Our learning rates are adjusted according to a cosine decaying policy~\cite{goyal2017accurate} and the initial learning rate is set to 0.005. The warm-up~\cite{goyal2017accurate} strategy is applied over the first 3 epochs, gradually increasing the learning rate linearly from 1e-6 to the initial value for the cosine schedule. The weight decay is set to 1e-5. The default image resolution is 320x320 during the training.

\subsection{Ablation Study}
\subsubsection{Model Architectures and Capacity}
We simply compare the models with different architectures and capacities.
We train some models with different architectures and capacities on the training split with some basic regularizations like dropout~\cite{srivastava2014dropout}. Table~\ref{table:model} shows the performances. We can see that the larger models perform the better, like most other deep learning tasks. And EfficientNet~\cite{tan2019efficientnet} surpasses ResNet~\cite{he2016deep} in the data-deficient setting. While Swin Transformer~\cite{liu2021swin} behaves badly. When the data is limited, the model needs more priors to learn from scratch on a small dataset, while Transformer, as is well known, lacks some of the inductive biases inherent to CNNs~\cite{dosovitskiy2021an}. Considering the better performance of the EfficientNet and the computing resource consumption, we use EfficientNet-b2 as the backbone in the latter experiments.

\subsubsection{Augmentations and Regularizations}
It is essential to enhance the generalization when data is limited. So we use some common data augmentation methods and regularization methods that are as follows: AutoAugment~\cite{cubuk2019autoaugment}, random erasing~\cite{zhong2020random}, dropout~\cite{srivastava2014dropout} with probability of 0.3, label smoothing~\cite{szegedy2016rethinking}, and using mixup~\cite{zhang2017mixup} with alpha of 0.5 or cutmix~\cite{yun2019cutmix} with alpha of 1.0 with a probability of 0.5. 
In addition, we find prolonged training epochs as 460 epochs can further improve the performance since learning effective features is difficult especially in a data-deficient setting.
The above methods enhance the generalization capacity of the model, leading to a quite good performance, and we pick it as our Strong Baseline. 
Results from Table~\ref{table:base} approve of the use of sufficient augmentations in improving the performance.

\begin{figure*}[tp]
  \centering
  \includegraphics[scale=0.4]{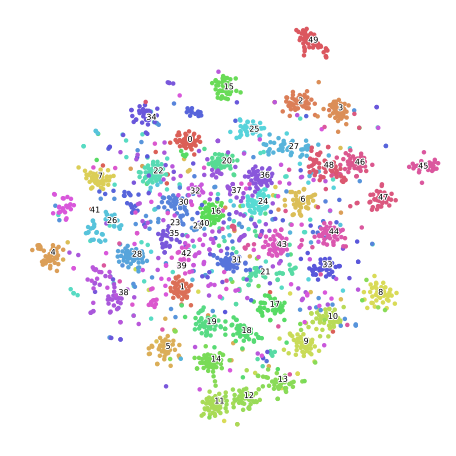}
  \hspace{0.5in}
  \includegraphics[scale=0.4]{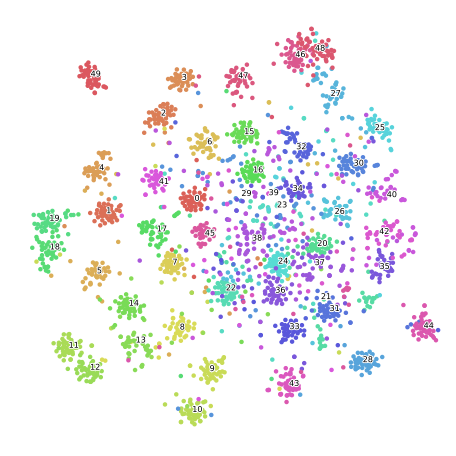}
  \caption{Visualization of latent features learned using baseline (left) and Attract-and-Repulse (right). 50 classes are randomly chosen from the 1000 classes of ImageNet and the same class are denoted by the same color. (Best viewed in color)}
  \label{fig:vis}
\end{figure*}

\subsubsection{Attract-and-Repulse Framework}
In this section, we perform ablation studies to demonstrate the effectiveness of the proposed Attract-and-Repulse framework over the VIPriors ImageNet training split with EfficientNet-b2.
As Table~\ref{table:ablation} shows, the SCE balance the learning of different classes to make the baseline better and the CR makes the biggest promotion by introducing the individual information. The Mean Teacher further enhance the performance and Auxiliary Classifier also have a slight refinement. As a result, the Attract-and-Repulse framework improving the top-1 accuracy from 58.77\% to 61.71\%. 

\begin{table}[t]
\begin{center}
\begin{tabular}{ll}
\hline\noalign{\smallskip}
EfficientNet-b2 & top-1 Acc. (\%) \\
\noalign{\smallskip}
\hline
Strong Baseline & 58.77 \\
\hline
AutoAugment $\rightarrow$ RandAugment & 57.25 \\
AutoAugment $\rightarrow$ Vanilla Aug & 55.21 \\
Mixup+Cutmix $\rightarrow$ Mixup &  57.75 \\
Mixup+Cutmix $\rightarrow$ Cutmix & 57.22 \\
Mixup+Cutmix $\rightarrow$ None & 54.76 \\
\hline
\end{tabular}
\caption{Ablation study on data augmentation for Strong Baseline.}
\vskip -0.2in
\label{table:base}
\end{center}
\end{table}
\begin{table}[t]
\begin{center}
\begin{tabular}{ll}
\hline\noalign{\smallskip}
EfficientNet-b2 & top-1 Acc. (\%) \\
\noalign{\smallskip}
\hline
Strong Baseline & 58.77 \\
+SCE & 59.50 \\
+Contrastive Regularization & 60.69 \\
+Mean Teacher & 61.46 \\
+Auxiliary Classiﬁer & 61.62 \\
+Auxiliary Fusion & 61.71 \\
\hline
\end{tabular}
\caption{Ablation Study of different components in Attract-and-Repulse framework.}
\vskip -0.2in
\label{table:ablation}
\end{center}
\end{table}

We further analyze the main components of the Attract-and-Repulse framework without postprocess like  Auxiliary Fusion. Figure~\ref{fig:ablation}(a) shows the influence of the weight ($\alpha$) for RCE in the SCE. It reaches a good performance when $\alpha=0.01$, while the performance decreases if keep increasing the $\alpha$. The effect of the weight ($\beta$) for CR is presented in Figure~\ref{fig:ablation}(b). The improving performance as $\beta$ increases demonstrates the CR is beneficial to the data-deficient learning. When $\beta$ is too large, the repulsion among instances suppresses the attraction of classes resulting in the degradation of performance. We also compare the CR with usual contrastive learning by features that add a projector after the backbone, results from Figure~\ref{fig:ablation}(b) show the performance of ``feature'' is much worse than CR. We further analyze the effect of $\delta$ in the CR. When $\delta=0$, the CR degrades to vanilla InfoNCE that only regards the two different views of the same images as positive pairs. As shown in Figure~\ref{fig:ablation}(d), we can see $\delta=0.005$ is optimal but the performance is changed slightly by $\delta$. Due to the large computing resource consumption, we only use a small batch size of 64 per GPU, so there are only a few sample pairs with a distance smaller than $\delta$. So the CR with small $\delta$ is similar because there are not many class-wise contrasts. While when $\delta$ is larger, many noisy samples are introduced into the contrasts resulting in the degradation. For the Mean Teacher, the influence of $\gamma$ is shown in Figure~\ref{fig:ablation}(c). The performance achieves the optimal when the weights for ground-truth and pseudo labels produced by the EMA model are balanceable.

\begin{table}[tp]
\centering
\tabcolsep=2pt 
\begin{tabular}[t]{ll}
\hline\noalign{\smallskip}
Method & top-1 Acc. (\%) \\
\noalign{\smallskip}
\hline
ResNet-50 Same-Conv~\cite{VIPriors2021} & 26.39 \\
ResNet-50 Full-Conv~\cite{VIPriors2021} & 31.16\\
Zhao's~\cite{zhao2020distilling} & 66.20 \\
Luo's~\cite{luo2020a} & 66.36 \\
Sun's~\cite{sun2020a} & 69.59 \\
\hline
Attract-and-Repulse (Ours) & \textbf{72.14} \\
\hline
\end{tabular}
\caption{Performance comparison with other state-of-the-art methods (single model version).}
\label{table:single}
\end{table}
\begin{table}[tp]
\centering
\begin{tabular}[t]{ll}
\hline\noalign{\smallskip}
Method & top-1 Acc. (\%) \\
\noalign{\smallskip}
\hline
Zhao's~\cite{zhao2020distilling} & 68.80 \\
Luo's~\cite{luo2020a} & 70.15 \\
Sun's~\cite{sun2020a} & 73.08 \\
\hline
Attract-and-Repulse (Ours) & \textbf{74.49} \\
\hline
\end{tabular}
\caption{Performance comparison with other state-of-the-art methods (model ensemble version).}
\label{table:ensemble}
\end{table}

\subsection{Visualization}

Figure~\ref{fig:vis} provides a t-SNE visualization~\cite{maaten2008visualizing} of the learned features for baseline (left) and Attract-and-Repulse (right). The figure illustrates how Attract-and-Repulse works. Due to the repulsion among individuals from different classes, the classes are more dispersed leading to a better generalization and performance. Although there also are repulsion among individuals in the same class, the features for the same class can still form cluster due to the attraction from the CE loss.

\subsection{Comparison with Other State-of-The-Arts}
For better performance in the VIProir competition, we combine the train and val split to train the model and randomly split a few samples for validation. And several other strategies and stronger backbone models are used for better performance (smaller batchsizes are used due to the larger computing resource consumption of stronger backbone).

We find that a larger resolution (448x448) can further boost the performance both on the training and the inference. And during the inference, TenCrop is utilized. After using these, we get an excellent performance (top-1 accuracy of 72.14\%) by a single model (EfficientNet-b7). Table~\ref{table:single} shows the comparison with other competing methods by using single model. Our framework is competitive and shows comparable results to current state-of-the-art methods.

Furthermore, experimental evidence shows that the ensemble method is usually much more accurate than a single model. We average the predictions of above methods in total of 16 models including EfficientNet-b5~\cite{tan2019efficientnet}, EfficientNet-b6, EfficientNet-b7, DSK-ResNeXt101~\cite{sun2020a}, ResNet-152~\cite{he2016deep}, SEResNet-152~\cite{hu2018squeeze}. Finally, we got the top-1 accuracy of 74.49\% on the testing set. We also compare with other state-of-the-art approaches by using model ensembling as shown in Table ~\ref{table:ensemble}, which demonstrates that our framework can surpass the previous state-of-the-art methods by up to 1.41\%, a significant improvement in the benchmark of ImageNet with only 8\% images.

\subsection{8\%(X) \emph{vs.} 100\%(X) \emph{vs.} 10\%(X)+90\%(U) ImageNet}
In this section, we compare the proposed Attract-and-Repulse framework using about 8\% labeled training data (``X'') with the popular fully-supervised methods using the entire ImageNet (100\% labeled training data, ``X''), as well as the popular semi-supervised methods using 10\% labeled training data (``X'') and 90\% unlabeled training data (``U''). Although there is a huge disparity of about 13\% top-1 accuracy between Attract-and-Repulse and the state-of-the-art fully-supervised methods, it needs to be mentioned that we merely use a tenth of ImageNet, and our result is approaching ResNet-50 with fully supervision. As for the semi-supervised with 10\% ImageNet, they use a bit more amount of labeled data and much more unlabeled data. It is very delightful that there is only a small gap with the SOTA results, and Attract-and-Repulse even surpasses some methods. We also train the NFNet-F6 and EfficientNet-B7 in the training setting of our Strong Baseline with 8\% ImageNet. Surprisingly, under the same condition, our Attract-and-Repulse surpasses them by a large margin higher than 2.72\%. (NFNet-F6 perfroms worse than EfficientNet-B7 perhaps because it is difficult to optimize the extremely large model with a small dataset.)

\begin{table}[tp]
\centering
\begin{tabular}[t]{lc}
\hline\noalign{\smallskip}
Method & top-1 Acc. (\%) \\
\noalign{\smallskip}
\hline
\emph{supervised with 100\%(X)} \\
VOLO-D5~\cite{yuan2021volo} & 87.1 \\
NFNet-F6~\cite{brock2021high} & 86.5 \\
SWIN-B~\cite{liu2021swin} & 84.5 \\
EfficientNet-B7~\cite{tan2019efficientnet} & 84.3 \\
SENet~\cite{hu2018squeeze} & 82.7 \\
ResNeXt-101~\cite{xie2017aggregated} & 80.9 \\
DenseNet-264~\cite{huang2017densely} & 77.9\\
ResNet-152~\cite{he2016deep} & 77.8 \\
FENet 1.375x + SE V2~\cite{chen2019all} & 76.5\\
ResNet-50~\cite{he2016deep} & 76.0 \\
\hline
\emph{semi-supervised with 10\%(X)+90\%(U)} \\
SimCLRv2~\cite{chen2020big} & 80.9 \\
TWIST~\cite{wang2021self} & 75.3 \\
CoMatch~\cite{li2020comatch} & 73.7 \\
FixMatch~\cite{sohn2020fixmatch} & 71.5 \\
\hline
\emph{supervised with 8\%(X)} \\
Strong Baseline (NFNet-F6) & 66.89 \\
Strong Baseline (EfficientNet-B7) & 69.42 \\
Attract-and-Repulse (EfficientNet-B7) & \textbf{72.14} \\
Attract-and-Repulse (Ensemble) & \textbf{74.49} \\
\hline
\end{tabular}
\caption{Performance comparison with fully-supervised learning and semi-supervised learning on ImageNet. ``X'' and ``U'' denote labeled and unlabeled data, respectively.}
\label{table:fullysemi}
\end{table}

Although these performance comparisons with fully-supervised with 100\% ImageNet or semi-supervised with 10\% ImageNet are somewhat ``unfair'', it can reflect the performance of our framework to some extent. Also, we can see that the improvement of the generalization algorithm still can not replace the role of tremendous training data.

\section{Conclusions}
It is of great challenge to learn from scratch on small-scale datasets efficiently since the model is difficult to learn the ambiguous concepts of categories or the model is easy to over-fit the data. We propose the Attract-and-Repulse framework to deal with the data-deficient image classification which consists of Contrastive Regularization, Symmetric Cross Entropy, and Mean Teacher. The Attract-and-Repulse mainly depends on the trade-off between the learning of instance-wise information and class-wise information, which can prevent the model from over-fitting the ambiguous classes. Ablation studies demonstrate that our framework is effective in data deficient learning. And together with other strategies, we achieve competitive performance in the VIPriors Image Classiﬁcation Challenge and surpass the previous state-of-the-art approaches. It is very delightful that Attract-and-Repulse achieves a competitive performance compared with the fully-supervised methods using the entire ImageNet and the semi-supervised methods with 10\% ImageNet. We hope the research direction of data deficient learning can take us close to the core of representation learning, and we hope our work can bring new inspirations to the community of representation learning.

\bigskip

%
%
\bibliographystyle{aaai22}
\bibliography{egbib}

\end{document}